# Machine Learning for UAV Propeller Fault Detection based on a Hybrid Data Generation Model

J. J. Tong[a], W. Zhang[a], F. Liao[b], C. F. Li[a], Y. F. Zhang[a]

[a]Department of Mechanical Engineering, National University of Singapore
[b]Temasek Laboratories, National University of Singapore

***Abstract –*** *This paper describes the development of an on-board data-driven system that can monitor and localize the fault in a quadrotor unmanned aerial vehicle (UAV) and at the same time, evaluate the degree of damage of the fault under real scenarios with interference information and without additional denoising procedures. To achieve offline training data generation, a hybrid approach is proposed for the development of a virtual data-generative model using a combination of data-driven models as well as well-established dynamic models that describe the kinematics of the UAV. To effectively represent the drop in performance of a faulty propeller, a variation of the deep neural network, known as the Long Short-Time Memory network (LSTM) is proposed. With the revolution per minute (RPM) of the propeller as input and depending on the fault condition of the propeller, the proposed propeller model estimates the resultant torque and thrust. Then, flight datasets of the UAV under "normal" conditions as well as various "fault" scenarios are generated via simulation using the developed data-generative model. Lastly, a fault classifier using a convolutional neural network structure (CNN) is proposed to identify as well as evaluate the degree of damage to the damaged propeller. The scope of this paper currently focuses on the identification of faulty propellers and classification of the fault level for quadrotor UAVs using their RPM as well as flight data. Doing so allows for early minor fault detection to prevent serious faults from occurring if the fault is left unrepaired. To further validate the workability of this approach outside of simulation, a real-flight test is conducted indoors. The real flight data is collected and a simulation to real (sim-real) test is conducted. Due to the imperfections in the build of our experimental UAV, a slight calibration approach to our simulation model is further proposed and the experimental results obtained show that our trained model can identify the location of propeller fault as well as the degree/type of damage. Currently, the diagnosis accuracy on the testing set is over 80%.*

(Corresponding author: Wei Zhang)

J.J. Tong and W. Zhang and Yunfeng Zhang are with the Department of Mechanical Engineering, National University of Singapore (email: tongjj@nus.edu.sg, weizhang@u.nus.edu, mpezyf@nus.edu.sg).

Fang Liao is with Temasek Laboratories, National University of Singapore. (email:tsllf@nus.edu.sg)

## I. Introduction

In recent years, unmanned aerial vehicles (UAVs) in the form of fixed-wing and multi-rotor are gaining more attention due to their significant usability and important application in many tasks such as surveillance [1], search and rescue [2-3], agriculture applications [4-5], as well as various military and security applications [6]. There are many types of UAVs available with various specialties. This paper focuses on the quadrotor UAVs that are often favored due to their small size, lightweight, and ease to control. These quadrotors refer to a type of UAV that each consists of two pairs of counter-rotating rotors and propellers located at the vertex of a square frame to ensure that it is dynamically balanced. Due to this unique configuration, damage or fault to any propellers could result in instability in the flight behavior of the quadrotor. In practice, "crack" and "bent" propellers are the most common faults encountered in the extensive use of quadrotor UAVs outdoors during operations. For minor "cracks" or "bent" present in the propeller, the quadrotor may be able to self-balance itself by compensating with higher RPM in its other propeller. This phenomenon can be often detected by a difference in stability and attitude when compared to healthy UAVs. In more serious cases, the quadrotor may not produce enough thrust to sustain its weight and eventually crash to the ground, potentially damaging the entire UAV's structure or causing injuries to its operator. Therefore, it is necessary to develop a robust onboard fault detection and identification system that can detect faults currently on the UAV during its early stage to avoid catastrophic failure. A system like this requires a better understanding of the behavior of a malfunctioning UAV which currently, requires the manual operation of a spoilt UAV. Such experiments expose the pilot to extreme danger as these malfunctioning UAVs exhibit unpredictable behavior. Thus, a data generative model that can replicate the behavior of malfunctioned UAVs is proposed.

## II. Related works

Research relating to UAV fault detection can generally be classified into three main categories, namely, hardware redundancy, model-based approach, and data-driven approach. In the research area of hardware redundancy, Panitsrisit et al. [7] proposed a hardware redundant system consisting of various inexact voters to detect faults in the elevator of the UAV by continuously comparing the states and functionalities with its sensors. Based on this idea,

Lieret et al. [8] further designed and implemented such a system for its flight control units. The proposed approach was evaluated on actual flights of a hex rotor.

Analytical model-based approaches for UAV fault detection focus on analytical models that adopt mathematical models and observable variables from the vehicle to determine the fault [9]. These methods mostly utilize a state estimation approach as well as parameter estimates. By setting a threshold value, fault in the UAV can be determined when the actual operation mode differs from its expected behavior. An early effort in this approach can be found in Chen et al. [10], who proposed an observation-based approach to determine fault in the quadrotor and verified the method using simulations. With regards to the rigid threshold value used by Chen et al., Avran et al. [11] proposed a nonlinear adaptive estimation that actively updates the threshold values to boost robustness and fault sensitivity in different scenarios. Rago et al. [12] and Zhang et al. [13] proposed a fault detection method to detect failures of sensors/actuators based on interactive multiple models. With the interactive multiple models, Zhong et al. [14] tackled the issue of multiple fault diagnosis for both actuators and the system of a quadrotor UAV. Even though model-based methods have good robustness and can diagnose unknown faults, the detection of system faults in quadrotors is not easily represented by analytical models due to their complicated structure. These model-based approaches are often not easy to implement in practical applications and lack scalability where the flight path of a UAV can be dramatically altered by wind or other external disturbances.

In recent years, data-driven approaches are getting more attention due to their robustness and reliability in fault detection. The fault detection based on a data-driven approach extracts various features from the original data and feeds them into the neural network model to obtain the fault detection results directly. The learning-based approach based on supervised learning requires experimental data containing the behavior of faulty quadrotors for training and labeling the fault cases [15]. In the case of an unlabeled fault, the result is predicted as a probability distribution based on the trained dataset. Guo et al. [16] proposed a fault detection approach based on hybrid feature models as well as an artificial neural network. By using a Short-time Fourier transform (STFT), the audio signals of the propeller can be converted into time-frequency spectrograms for fault detection. Subsequently, more robust models based on long short-term memory (LSTM) models [17] are introduced to more accurately detect the fault in various quadrotors. Liu et al. [18], on the other hand, proposed a detection approach based on a convolutional neural network coupled with a transfer learning method. For the detection of sensor faults, Chen et al. [19] proposed a wavelet packet deposition (WPD) to extract the energy entropy as a feature to train a generic backpropagation (BP) neural network. Similarly, Xiao et al. [20] extracts the energy entropy and proposed an observer method based on the BP algorithm to detect sensor faults in real-time. Although learning-based approaches seem to be the ultimate solution in accurately detecting UAV faults, the growth in this approach is unfortunately plateaued by the absence of a reliable data generation approach to capture fault datasets. In [21], a dataset containing "fault condition" are collected by artificially destroying the propeller in a small area. By inducing such damage to the propeller, the output thrust and torque of the propeller will change, thus affecting the flight attitude of the whole UAV. However, due to safety reasons, only minor damage could be induced to the propeller so that the UAV will not suddenly crash. The dataset generated in this manner is thus very limited and does not represent scenarios where there could be more serious damage to its propellers. To rectify these limitations, a reliable and efficient workflow in data generation of fault detection datasets is needed.

In this paper, a data-driven approach is proposed for the development of the system where flight signals of the UAV captured onboard are used as inputs for classification. As data-driven systems for fault detection requires a large amount of dataset consisting of various quadrotor fault scenario, one feasible solution is to simulate the output signals using a hybrid data generation model of the UAV developed from existing open-source simulation [23]. The roadmap for developing a UAV fault diagnosis system is proposed as shown in Fig. 1.

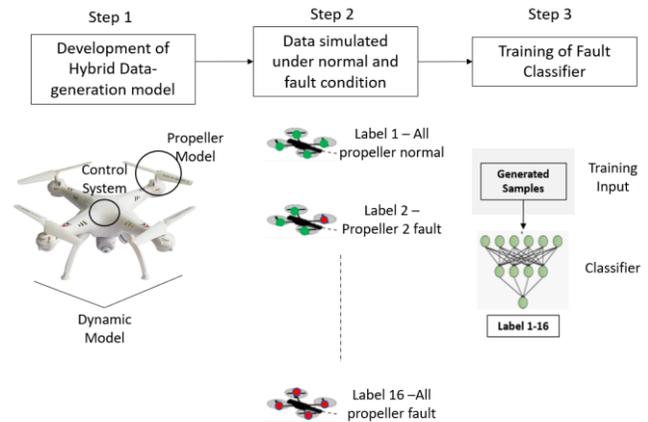

*Figure 1. The overall framework of the proposed 3-step UAV fault diagnosis approach.*

As shown in Fig. 1, Step 1 involves the development of the hybrid data generation model for the adopted quadrotor. In Step 2, the data generation model is used to generate data under both normal and specified fault conditions, in which 16 categorical labels are defined to specify the position and condition of the 4 propellers, respectively. With the generated training samples and their labels, Step 3 describes the formulation of a learning-based model to classify and identify the location and severity of the propeller fault.

The novel contributions of this paper are summarized as follows:

1. An end-to-end quadrotor's fault detection approach based on a novel hybrid data-generative model is presented. To the best of our knowledge, this paper presents the first data-generative model based on a data-driven model trained using realistic loadcell experiments to accurately capture the behavior of a faulty propeller when mounted on a quadrotor.
2. Existing simulation models tend to omit the effects of the imperfect build of the quadrotor structure such as a non-centered center of gravity (CG). From our experience in dealing with a quadrotor, non-centered CG is a common occurrence and failure to account for this will lead to more inaccurate results. Thus in this aspect, we present a logical approach to account for such imbalances, making the classifier more accurate.
3. Fault in a propeller comes in different forms which are often tedious to represent with a conventional method. To better represent the drop in performance of the faulty propeller over time, an LSTM network is proposed.
4. Many feature-based approaches shrink the input data into lower dimensional space and often omit important fault characteristics information. To prevent this limitation, a non-feature-based approach based on a two-dimensional convolutional neural network is proposed to adaptively extract features from the original flight data to detect the location and severity of the propeller fault.

The rest of the paper is structured as follows. Section III describes the development of the hybrid data generation model combining a data-driven model and an analytical dynamic model to generate flight data of the quadrotor. Section IV presents the data generation process using a developed data generative model under both normal and fault conditions. In this section, the training and testing results of the proposed convolutional neural network (CNN) fault classifier are further discussed, ending with the conclusion and future work in section V.

### III. PROPOSED METHODS

#### A. Overview

The end-to-end fault diagnosis approach using flight data consists of two main components, namely, the data generative model as well as the fault detection model. This section introduces the design principles and architecture of the proposed data generative model. Fig. 2 shows the workflow of the mentioned approach in which the dynamic model (Top) generates the flight data, and a fault classifier (Bottom) detects the type of propeller fault as well as the location of its fault.

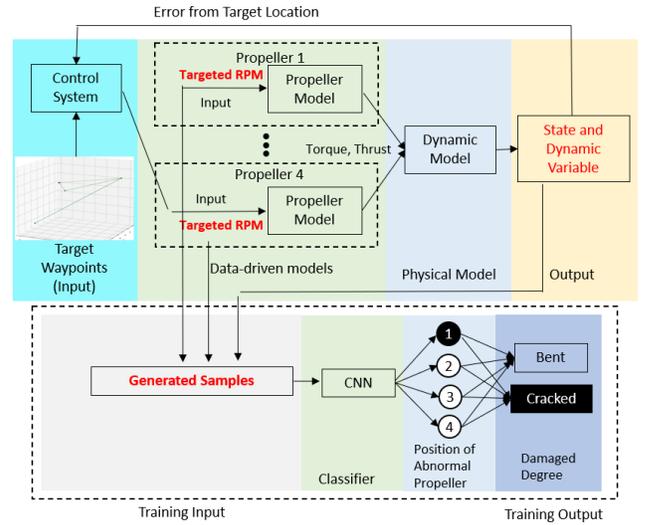

*Figure 2: The framework of the data generative model and fault classifier of the UAV. The text marked in red means the corresponding signal is known or can be measured during flight.*

#### B. Data-generative UAV model

The proposed quadrotor UAV data generation model is shown in Fig. 3 and consists of three subsystems, i.e., the Control system, Propeller Model, and Dynamic Model. As the damage in the Control system and BLDC motor may happen abruptly and is unpredictable, in this paper, we assume that this system is not damaged and will only include in our future research. The Dynamic Model which is well described by the physical models is used to compute the flight data based on the input RPM of the four individual propellers. In this subsection, we will first describe the development of the Propeller Model. Subsequently, we will discuss how we incorporate the propeller model into our simulation model to accurately represent cases of a quadrotor with a damaged propeller.

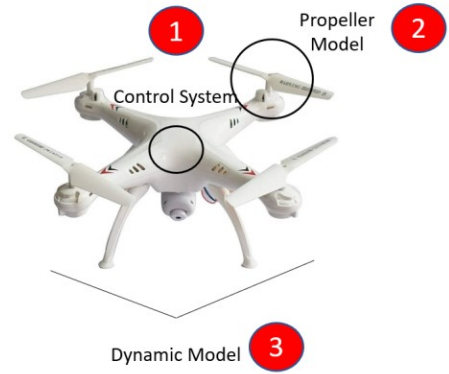

*Figure 3. Components of the proposed data generative model*

The workflow of the proposed data generative model is shown in Fig. 2 (Top). The input to the simulation system is a set of targeted waypoints where the control system will first compute four target RPMs for each propeller in the quadrotor. Subsequently, the four RPMs are passed to the Propeller Model to generate the torque and thrust, which are then used to compute the kinematics of the quadrotor body using the theoretical Dynamic Model. The position and velocity of the quadrotor (*x*, *y*, *z*, roll, pitch, yaw, *Vx*, *Vy*, and *Vz*) at the current timestep are recorded and the position of the quadrotor is passed back through the feedback loop so that the control system can compute the target RPM for the next timestep to drive the quadrotor closer to the target point. The details of the individual subsystems are described in the following sub-sections.

1. *Autopilot Controller*

The controller used in the data generative model is the PX4 autopilot controller system, which is an open-source flight control software for drones and other unmanned vehicles. In our data generative model, the target waypoints are fed into the system as inputs, and the autopilot control system computes a series of target RPM and the required control signals in the form of an ESC signal to guide the UAV toward the targets. To achieve this, PX4 uses sensors to determine vehicle states that are needed for both stabilizations and to enable autonomous control. Some common examples include a gyroscope, accelerometer, magnetometer (compass), and barometer.

2. *Propeller Model based on LSTM network*

Conventionally, under normal conditions, the thrust (*f*) and torque (*τ*) generated by a propeller can be computed using Eqs. (2) and (3),

$$f_i = k_f * \omega_i^2 \quad (2)$$

$$\tau_i = k_\tau * \omega_i^2 \quad (3)$$

where *i* is the propeller number, $\omega$ is the rotational speed (RPM). $k_f$ is lift constant and $k_\tau$ the drag constant, both of which can be experimentally determined.

However, when the propeller is bent or cracked, the linear relationship shown in Eqs. (2) and (3) may no longer hold. An effective way to account for such non-linearity is to use a learning-based approach, in this case, an LSTM network. The input to the network is RPM, and the outputs are thrust and torque, respectively. Depending on the condition of the propeller, three types of propeller models, representing the condition of the propeller ("Normal", "Bent", "Cracked"), are trained as shown in Fig. 5.

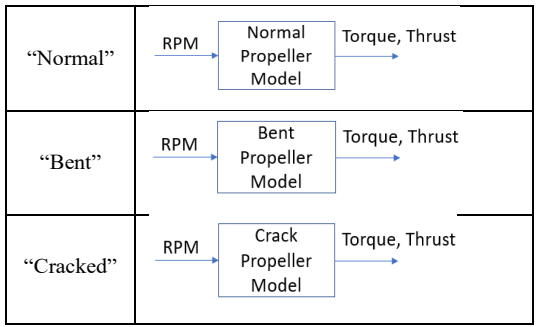

Figure 5. Three propeller models, each representing a specific propeller condition

As LSTM has been proven to achieve good results in time series problems and is capable of learning long-term dependencies, we have chosen LSTM as the basic regression prediction model for the Propeller Model.

Recurrent neural networks (RNNs) have layers of repeating modules of a neural network shown. In standard RNNs, this repeating module will have a very simple structure, such as a single tanh layer. LSTM, on the other hand, has four neural network layers, each interacting in a very special way. The hidden node of the LSTM layer is a memory cell shown in Fig. 6. The basic cells are composed of three gates, the input gate, forget gate, and the output gate [27].

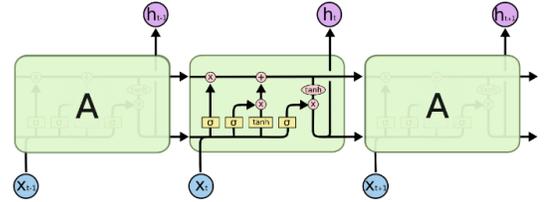

The repeating module in an LSTM contains four interacting layers.
Figure 6. Repeating modules in LSTM

When a *d*-dimensional input $x_t \in R^{d*1}$ arrives at time *t*, the cell is updated, and the new information is recorded. Assuming there are *k* cells in one LSTM layer, and the cell states and hidden states at time $t - 1$ are defined as $c_{t-1} \in R^{k*1}$ and $h_{t-1} \in R^{k*1}$, respectively. The new cell states $c_t$ and hidden states $h_t$ at time *t* is updated by the following formulas:

$$f_t = sigmoid(w_f \cdot [h_{t-1}, x_t] + b_f) \quad (4)$$

$$i_t = sigmoid(w_i \cdot [h_{t-1}, x_t] + b_i) \quad (5)$$

$$\tilde{c}_t = tanh(w_c \cdot [h_{t-1}, x_t] + b_c) \quad (6)$$

$$o_t = sigmoid(w_o \cdot [h_{t-1}, x_t] + b_o) \quad (7)$$

$$c_t = f_t * c_{t-1} + i_t * \tilde{c}_t \quad (8)$$

$$h_t = o_t * \tanh(c_t) \quad (9)$$

where $f_t$, $i_t$, and $o_t$ are the outputs of forget gate, input gate, and output gate, respectively. $\tilde{c}_t$ is an intermediate variable used to update $c_t$. $w_f \in R^{k(d+k)}$, $w_i \in R^{k(d+k)}$, $w_o \in R^{k(d+k)}$, and $w_c \in R^{k(d+k)}$ represent the weight matrices. $b_f \in R^{k*1}$, $b_i \in R^{k*1}$, $b_o \in R^{k*1}$, and $b_c \in R^{k*1}$ are biases. Besides, sigmoid and tanh refer to the sigmoid activation function and hyperbolic tangent activation function, respectively. The symbol $*$ denotes the Hadamard product.

After calculating from (4) to (9), the LSTM memory cells remember the information from the beginning to the moment *t*. $h_t$ is the final output of these cells at time *t*. As the input sequence arrives, these cells' outputs predict values continually, which, in turn, can form an output sequence, or we can only use the final output as a sequence-to-one prediction. To predict the thrust and torque of a propeller given RPM as input, the stacked LSTM model [27] is used here. Based on the function of a single LSTM layer, the stacked LSTM regression model is shown in Fig. 7. There are two LSTM layers and two dense layers in this model. The input shape is determined by the dimension of the input vector, in this case, a size of 1. The first LSTM layer works in the sequential mode, and the second LSTM layer only outputs one point for each input sequence. The dense layer is a fully-connected layer with a linear kernel. To achieve single-step prediction, input variables $x \in R^{N*d}$ and target variable $Y \in R^{N*1}$ need to be reconstructed with a sliding window. The length of the window is $L$, which means the past $L$ historical samples are used to predict the monitored parameter at the next moment. After the reconstruction, the new input samples and corresponding outputs are obtained as follows:

$$X_R = \{x_1, x_2, \ldots \ldots, x_{N-L}\}, \quad x_i \in R^{L*d}$$

$$Y_R = \{y_1, y_2, \ldots \ldots, y_{N-L}\}, \quad y_i \in R^{L*d} \quad (10)$$

The stacked LSTM regression model needs to learn the mapping function $f_{LSTM}(.)$, which is defined as

$$\tilde{Y}_R = f_{LSTM}(X_R) \quad (11)$$

where $\tilde{Y}_R$ is the prediction of $Y_R$. The mean square error (MSE) is used as loss training. The proposed propeller model is shown in Fig. 8.

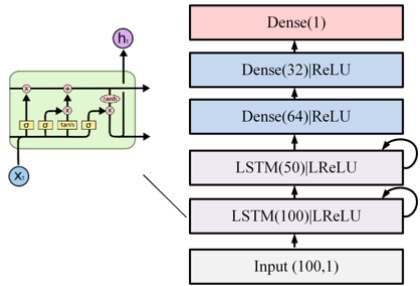

Figure 7. Propeller Model's network structure

### 3. *Dynamic Model to describe the Kinematics of the quadrotor*

With the torque and thrust of the individual propeller generated by the Propeller Model, the kinematics of the quadrotor can then be computed using well-established physical models [28]. To induce motion in a quadrotor, control mechanisms by roll, pitch, and yaw is adopted. These are represented by the angle of rotation around the center of the quadrotor's body.

In general, to track the altitude of the quadrotor, a two-coordinate system is usually required (see Fig. 8). The inertial coordinate system describes the coordinate system fixed to the earth and is independent of the quadrotor motion while the body frame system is attached to the quadrotor's body at its center of gravity. The angular difference between the two coordinates describes the behavior of the quadrotor attitude in space.

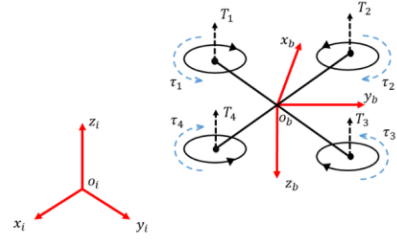

Figure 8. The inertia and body coordinate frame of the quadrotor

To describe the motion of the quadrotor, there are 12 states of the quadrotor as shown below.

$$X = [\emptyset, \theta, \psi, \dot{\emptyset}, \dot{\theta}, \dot{\psi}, X, Y, Z, \dot{X}, \dot{Y}, \dot{Z}] \quad (11)$$

The angle of roll, pitch, and yaw are represented with $\emptyset, \theta, \psi$ while its angular velocity is represented as $\dot{\emptyset}, \dot{\theta}, \dot{\psi}$. These states represent the relationship between the quadrotor and the inertia coordinate system. The next six states represent the physical relationship of the quadrotor's physical location within the earth-fixed system and are denoted as $X$, $Y$ and $Z$. In addition, the quadrotor velocity along these axes is denoted as $\dot{X}, \dot{Y}, \dot{Z}$ respectively.

In essence, the movement of a quadrotor is induced by the difference in torque and thrust of each of its four propellers by forcing a change around the pitch, roll, and yaw angle. The Dynamic Model of the quadrotor consists of the rotational subsystem that represents the roll, pitch, and yaw angle, and the translational subsystem that represents the $X$, $Y$, and $Z$ positions. The full derivation can be found in [28]. Applying the Newton-Euler equation to the quadrotor body results in the equations of motion, (12) and (13), which summarize the kinematic of the quadrotor's body.

$$m\ddot{x} = \begin{bmatrix} 0 \\ 0 \\ -mg \end{bmatrix} + RT_B + F_D \quad (12)$$

$$\dot{\omega} = \begin{bmatrix} \tau_\emptyset I - I_{xx} \\ \tau_\emptyset I - I_{yy} \\ \tau_\emptyset I - I_{zz} \end{bmatrix} - \begin{bmatrix} \frac{I_{yy} - I_{zz}}{I_{xx}} \omega_y \omega_z \\ \frac{I_{zz} - I_{xx}}{I_{yy}} \omega_x \omega_z \\ \frac{I_{xx} - I_{yy}}{I_{zz}} \omega_x \omega_y \end{bmatrix} \quad (13)$$

*3.1 Simulation model Adjustment*

To apply real-flight fault diagnosis, we observed that the CNN model trained using the original simulation model performed poorly in real flight with the diagnosis accuracy dropping below 50%. The main cause is that in the original simulation model, the CG of the quadrotor is centered, while the CG of the experimental UAV is off-center. In the simulation, we follow an ideal dynamic model of the quadrotor, i.e., if given four propellers with the same rotation speed, it should be in an upright hovering position with angular velocity and acceleration equal to zero.

However, in the real world, this behavior is often not followed because of the following factors: 1) the center of gravity that is off-centered; 2) the imperfect condition of the mounted propeller or motor; 3) the inaccuracy of flight sensors as well as the imperfect build of the quadrotor's structure. As shown in Fig. 9a, all motors' RPMs in the simulation are near the same at the hovering state, while all motors' RPMs (Fig. 9b) in the real flight are not the same at the hovering state. Therefore, for our simulation to accurately replicate real flight data more accurately, we need to calibrate our simulation model to account for this phenomenon. As shown in figure 9b below, the average speed of motor 4 is higher than motor 1 at a hovering state, which means that, on average, motor 4 must rotate faster to keep the quadrotor in a stable position. One way to simulate such behavior is to assume motor 4 is weaker than motor 1 such that it is required to rotate at a faster rate to produce similar torque and thrust. From the generated loadcell data in our flight test, the unbalanced ratio ($U_r$) can be computed using the average RPM ($\overline{\omega_i}$) of each respective motor over the baseline motor 1 ($\overline{\omega_1}$), i.e.,

$$U_r^i = \frac{\overline{\omega_i}}{\overline{\omega_1}} \quad (14)$$

$$U_r(\omega_i) = 1 + \frac{\omega_i}{\max(\omega_i)} U_r^i \quad (15)$$

With the computed unbalanced ratio for each respective RPM, the adjusted torque and thrust generated by each propeller are as follows,

$$F_i = \frac{\omega_i^2}{U_r^2(\omega_i)} \quad (16)$$

$$\tau_i = k_\tau \frac{\omega_i^2}{U_r^2(\omega_i)} \quad (17)$$

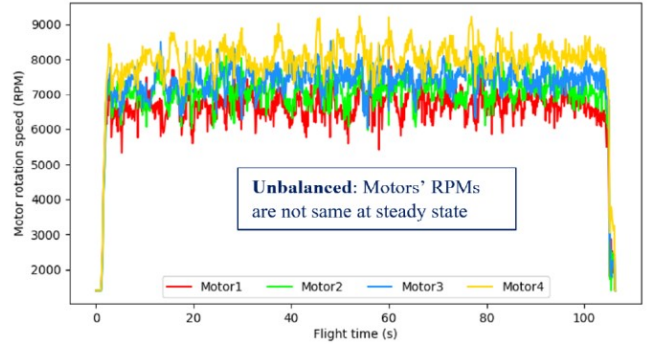

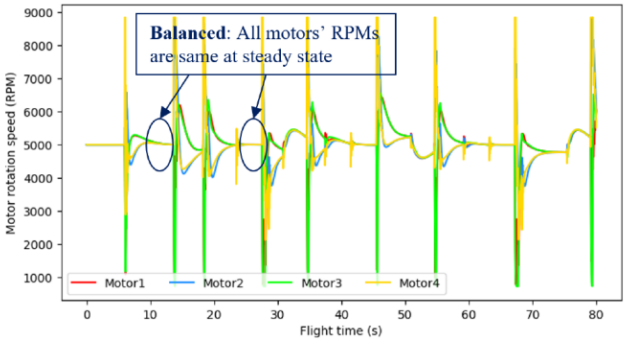

Figure 9a. RPM vs Time curve for simulation (top) and, 9b real-flight (bottom)

In this manner, the simulated model will exhibit similar unbalanced behavior as our real-world quadrotor.

### C. CNN-based fault classifier

With the trained Propeller Model and the Dynamic Model defined, we can now conduct simulation flight runs to generate the output signals (see Table 1) of the UAV under normal and fault conditions. As the output signal needs to be obtainable using onboard sensors, not all output signals can be used for training of fault classifier.

Table 1: Parameters simulated by the data generative model

| Symbol | Description |
|---|---|
| $t$ | Time |
| $X, Y, Z$ | Position |
| $\dot{X}, \dot{Y}, \dot{Z}$ | Velocity along $X$, $Y$, and $Z$ directions |
| $\phi, \theta, \psi$ | Roll, Pitch, Yaw |
| $\dot{\phi}, \dot{\theta}, \dot{\psi}$ | Roll Rate, Pitch Rate, Yaw Rate |
| $\omega_1, \omega_2, \omega_3, \omega_4$ | RPMs of Propellers 1, 2, 3, 4 |
| $f_1, f_2, f_3, f_4$ | Thrusts of Propellers 1, 2, 3, 4 |
| $\tau_1, \tau_2, \tau_3, \tau_4$ | Torques of Propellers 1, 2, 3, 4 |
| $\delta_X, \delta_Y, \delta_Z$ | Errors in $X$, $Y$, and $Z$ from targeted waypoint |

The chosen training parameter is the quadrotor's angular acceleration as well as the respective four RPM of the motor. Quadrotor angular acceleration ($\ddot{\phi}, \ddot{\theta}, \ddot{\psi}$) can be directly measured by the onboard sensor. On a real flight, however, the motor velocities cannot be directly measured. Thus, based on our loadcell experiment, we derived the relationship between the control signal ESC and the Rpm of the respective propeller as follows,

$$RPM_i = -0.0062(ESC)^2 + 29.37(ESC) - 22992$$

In this segment, propeller faults are classified into two categories (fault/normal). Based on the location of the malfunctioning propeller of the quadrotor UAV, 16 labeled scenarios are categorized as shown in Fig. 10. For example, label 1 represents the case where all propellers are normal, label 2: propeller 1 is faulty and the other 3 are normal, and label 16: all 4 propellers are faulty.

A more detailed illustration of how we incorporate our trained propeller model into our simulation model is described in the following examples. Below show the steps for generating datasets by conducting flight simulations under the different labeled categories.

Data collection under label 1 (refer to Fig. 10)

[1] For all motors, choose the "Normal" Motor Model.
[2] Choose the "Normal" Propeller Model for all propellers.
[3] Feed the target waypoints to the UAV model, simulate the corresponding control signal, voltage signal, and output signal and record the values (refer to Table 1) at their corresponding timesteps.
[4] Add the control signal, the voltage, and the output signal into the dataset under "label 1".

Data collection under label 5 (refer to Fig. 10)

[1] For all motors, choose the "Normal" Motor Model.
[2] Choose the "Normal" Propeller Model for propellers 1-3 and the "Fault" Propeller Model for propeller 4.
[3] Feed the target waypoints to the UAV model, simulate the corresponding control signal, voltage signal, and output signal and record the values at their corresponding timestamp.
[4] Add the control signal, the voltage, and the output signal into the dataset under "label 5".

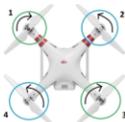

Figure 10. The proposed 16 labeled categories and their corresponding fault type.

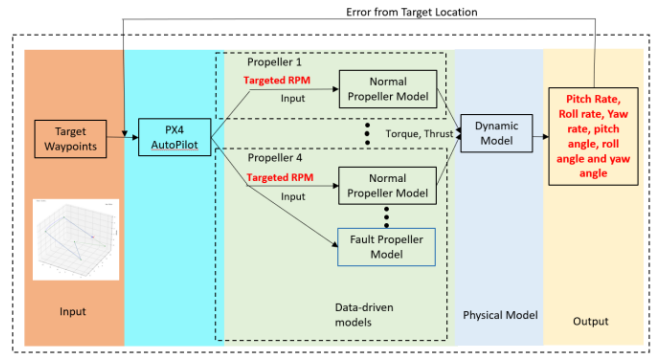

Figure 11. Data generation using data generative model simulation for label 5.

We can repeat this procedure to generate labeled datasets under all 16 conditions. From the list of parameters shown in Table 1, 10 variables (target RPMs, pitch, roll, yaw, pitch rate, yaw rate, and roll rate) are chosen as input for the training of the classifier. The training dataset under each label thus consists of the 10 variables shown in Fig. 12.

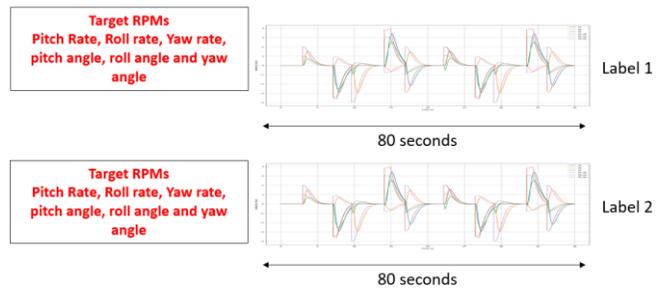

Figure 12. Generated dataset and their corresponding label

With the training samples collected from the simulation flight runs, we then train a CNN as a classifier for fault diagnosis. The overall framework of our fault diagnosis model is given in Fig. 13, in which the output layer consists of 16 labels, specifying the condition of the 4 propellers.

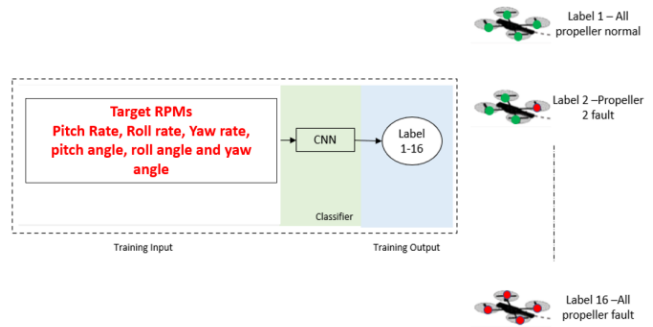

Figure 13. The framework of our fault classifier.

The network architecture is shown in Fig. 14. For each set of 80-second data, the timestep is set as 0.05 seconds and one training sample consists of 100 timesteps. Thus, this resulted in a training sample size of 100 x 8 matrix. The best way to ensure the trained network's generalization ability is by increasing the number of training samples. One simple way to do so is by overlapping the sampling data such that two consecutive training sample has 99 overlapped sampling data or timesteps. In this way, the number of training samples is largely increased. Furthermore, to increase the receptive field and pick up important features from the data, this input will first pass through a fixed kernel size of 3x3, and the number of convolution kernels is 32. The number of convolution kernels in each convolution kernel is doubled that of the previous module. In addition, a max-pooling layer is adopted after each convolutional layer to reduce parameters, keeping the important ones. The resulting parameters are flattened into one-dimensional vectors and input into the fully connected layer. Finally, the fault diagnosis results are obtained through the SoftMax layer. Each fault type corresponds to each output from the SoftMax layer respectively.

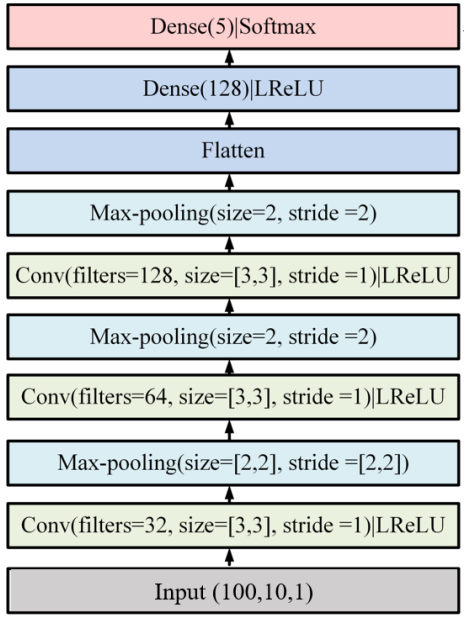

*Figure 14. Fault Classifier's network architecture.*

## IV. EXPERIMENTS AND RESULTS

### A. Data Collection and Training of Propeller Model

To build a learning-based Propeller Model, we first collect the training dataset with induced artificial damage to the propeller (see Fig. 15). We start by constructing an isolated system known as the loadcell as shown in Fig. 16 for data collection. In this setup, a single propeller is controlled by an isolated control unit with a constant voltage and the in-built sensors to allow for various measurements. These measurements include the acceleration vector, torque, thrust, RPM, vibrations as well as efficiency of the propeller. To train our propeller model, we use the RPM, Thrust, and Torque values. The experimental dataset is generated with the ESC signals varied from 1000 to 2000, which is the operating limit of the system.

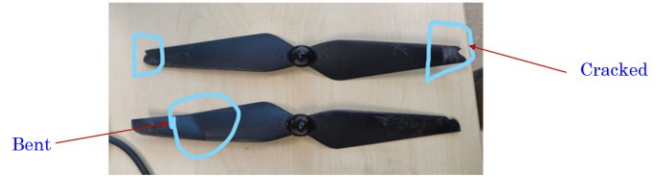

*Figure 15. Artificially induced damage to the quadrotor's propeller.*

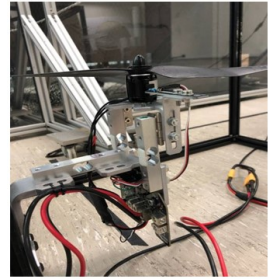

*Figure 16: Experimental data generation using a loadcell*

For each labeled case ("Normal", "Bent", and "Crack"), the experiment is conducted for approximately 5 minutes with a sampling period of 25ms. The torque and thrust measured under the "Normal" condition are shown in Fig. 17, respectively. As the experiment is conducted by ramping the control signal between 1000 to 2000, the peak in Fig. 17 corresponds to the case where the control signal is set at its highest while the trough corresponds to the control signal at its lowest. To ensure the LSTM model does not overfit, the training set will thus consist of the first 80% of the duration of the dataset while the remaining 20% is used as a test set.

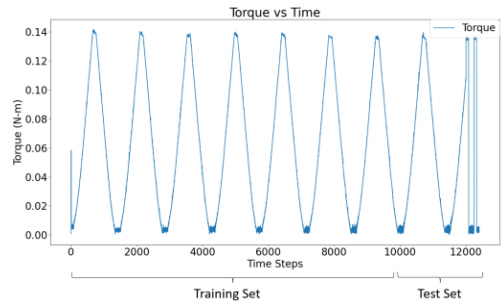

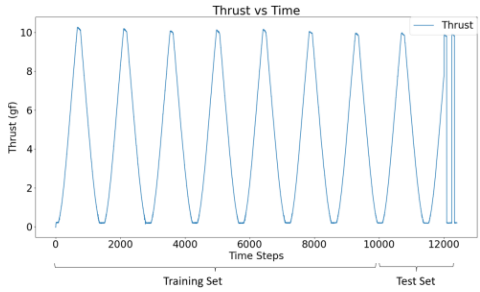

Figure 17. Normalized torque and thrust of the experimental dataset ("Normal").

Three propeller models, namely "Normal", "Bent", and "Cracked", have been trained using the respective datasets. Each model takes the RPM as input and computes the torque and thrust under its respective health condition. In the training process, the batch size is defined to be 32, and the learning rate is set at 0.01. The training ceases upon reaching convergence.

The "Normal" Propeller Model
Fig. 18 and Fig. 19 show the testing results of the "Normal" Propeller Model in which the blue curve represents the ground truth while the red curve represents the network predicted results. We can see that the network can predict the torque and thrust quite accurately for the propeller operating under the "Normal" condition with most of its error occurring at ESC = 2000, which is the operating limit of the motor. This does not cause much of an issue as the UAV is rarely required to operate at maximum power. The average error rate measured for torque is 2.29% while the average error rate measured for thrust is 0.613%.

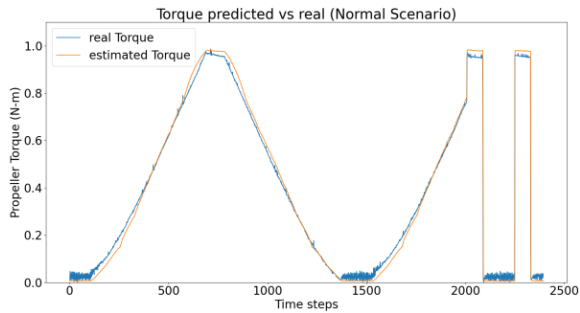

Figure 18. Testing result of the propeller model under a "normal" scenario (Torque)

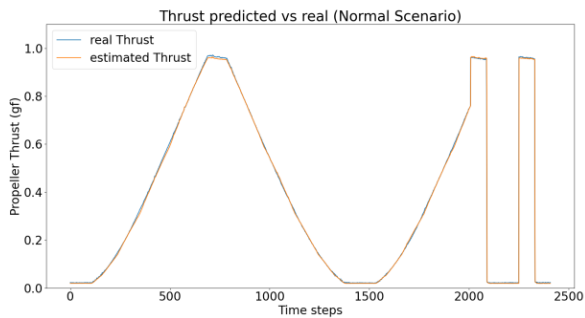

Figure 19. Testing result of the propeller model under "normal" scenario (Thrust)

The "Bent" Propeller Model
In the previous sub-section, we have shown that the normal propeller model can model the normal condition of the propeller quite accurately. Fig. 20 and Fig. 21 show the testing results of the "Bent" propeller model in which the blue curve represents the ground truth while the red curve represents the network predicted results. From Fig. 20, we can see that the torque prediction outperforms that of the normal condition achieving an average error rate of 1.37% while the average error rate for thrust is measured at 0.689%.

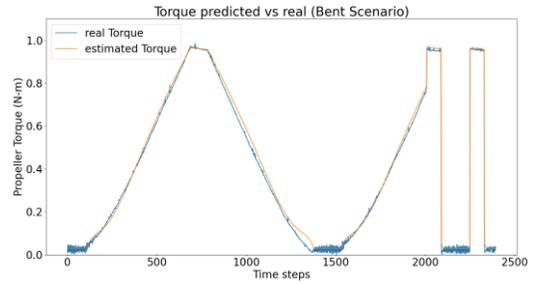

Figure 20. Testing result of the propeller model under "bent" scenario (Torque)

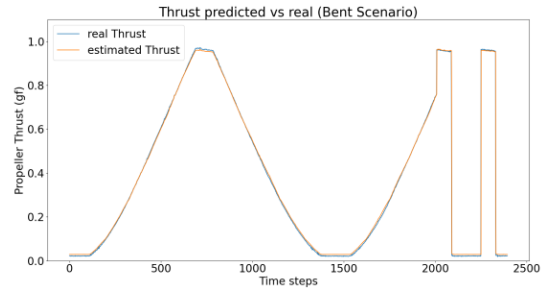

Figure 21. Testing result of the propeller model under "bent" scenario (Thrust)

The "Crack" Propeller model
Fig. 22 and Fig. 23 show the testing results of the "Cracked" propeller model in which the blue curve represents the ground truth while the red curve represents the network predicted results. As compared to the "Normal" and "Bent" condition, the "Cracked" propeller model did not perform as well as the previous fault conditions. In this case, the model slightly overestimated the torque during timestep 600-1300. The average error percentage for torque is measured at 2.47% while the average error percentage for thrust is measured at 4.27%.

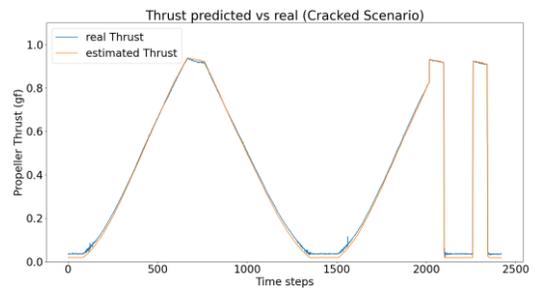

Figure 22. Testing result of the propeller model under "cracked" scenario (Thrust)

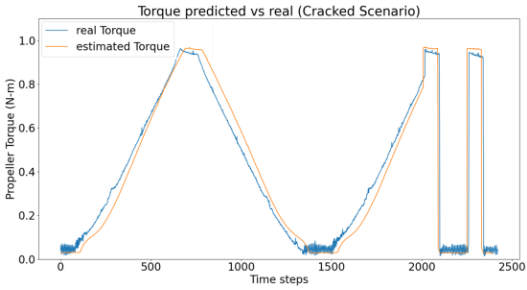

*Figure 23. Testing result of the propeller model under "cracked" scenario (Torque)*

### B. Data Collection and Training of Fault Classifier

The training datasets are generated by conducting flight runs in the simulation environment under normal scenarios as well as scenarios where one or more of its propellers are faulty. For each flight run, 5 waypoints are placed randomly in the 3D space. The overall flight duration is set at 80 seconds, where the quadrotor will approach the targeted waypoints starting from waypoints 1 to 5. If the quadrotor completes the designated route before the timer reaches 80 seconds, it will repeat its task until the end of 80 seconds.

The detailed parameters of the UAV model are listed in the table below. The mass of the quadrotor used in data collection is 1.2 kg with an arm length of 16cm. The torque and thrust coefficient is $1.076*10^{-5}$ N/rpm$^2$ and $1.632*10^{-7}$ Nm/RPM$^2$ respectively. The sampling period is set at 0.05 seconds, resulting in 1600 data points (80/0.05). As mentioned earlier, to increase the number of training samples, we use a window size of 100 and a hop length of 1 resulting in the total training set equal to 15802 samples.

Table 4.1: Parameters of UAV model II

| Parameters | UAV model |
|---|---|
| $L$ | 0.16 |
| $K_F$ | $1.076 * 10^{-5}$ |
| $K_t$ | $1.632 * 10^{-3}$ |
| $I_{xx}$ | 0.0123 |
| $I_{yy}$ | 0.0123 |
| $I_{zz}$ | 0.0224 |

Table 4.2: Parameters of UAV model II

| Dataset Type | Dataset Name | Weight (Kg) | Waypoints Label |
|---|---|---|---|
| *Training* | A | 1.5 | A |
| *Testing (Simulation)* | B | 1.5 | B |
|  | C | 1.95 | A |
| *Testing (Real-world)* | D | 2.025 | Nil |

For each set of 80-second data, the target RPM, voltage, roll, pitch, yaw, roll rate, pitch rate, and yaw rate are used as input to the network. This results in a dimension of [10, 100] for the training sample, each corresponding to one category, 1 to 16. An example of one data sample is shown in Fig. 25.

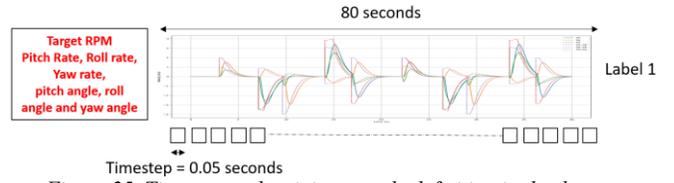

*Figure 25. Timestep and training sample definition in the dataset*

To ensure that the trained classifier does not overfit, 80% of the samples are used for training and 20% are used for testing the network. The datasets are fed into the classifier with network structure illustrated in Fig. 14. The classifier obtains a training accuracy of 100% and a testing accuracy of 99.97%. The overall summary of the classification accuracy is shown in Fig.26.

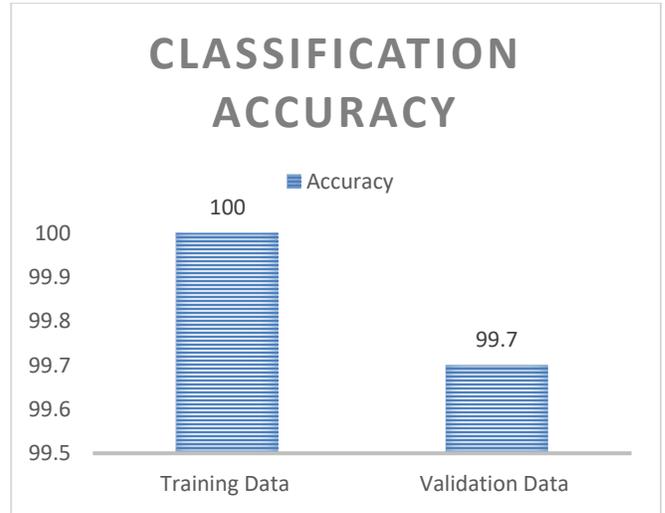

*Figure 26. Classification Accuracy of Fault Classifier*

## C. Results and comparison study

Furthermore, to test the performance of the trained classifier under different operating conditions, datasets are collected from flight runs following different targeted waypoints and payloads. Fig. 26 shows the flight trajectories used for data collection: (a) same waypoints with training dataset but the different payload, (b) different waypoints but same payload.

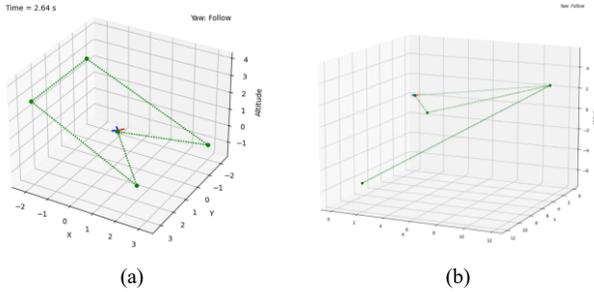

(a) (b)

*Figure 27. Testing data collection of two flight runs*

### Testing Dataset B: Different target waypoints

Using the waypoint shown in Fig. 27b, datasets under similar fault scenarios are generated while keeping the UAV weight constant at 1.5kg the testing accuracy obtained in this test is 85.66%. The corresponding confusion matrix is illustrated in Fig. 28. As the training datasets only consist of the quadrotor flying under waypoints label A, the classification model tends to overfit which explains why the testing accuracy under Dataset B is significantly lower than the validation accuracy. This issue can be resolved by introducing more waypoints in the training dataset so that the classification is accustomed to different flight paths of the quadrotor. To illustrate if the classification model can accurately detect faulty quadrotors, the confusion matrix is broken down into 2 categories, namely fault or normal. Another metric known as recall and precision is further computed. The precision is computed to be 96.5% while the recall is computed to be 99.8%.

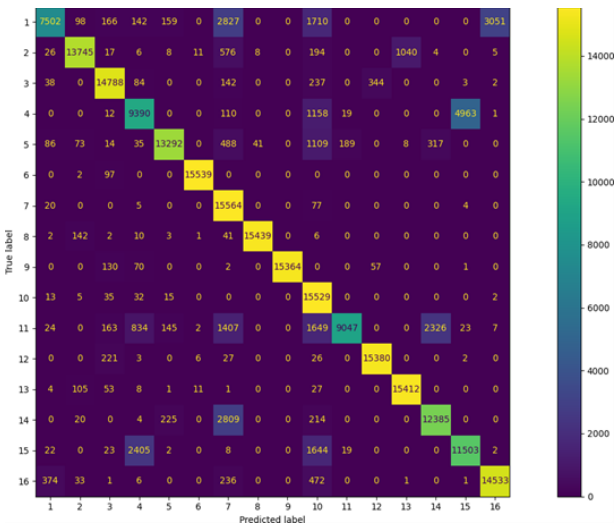

*Figure 28. Confusion Matrix of testing results (different flight path)*

### Test set 2: Increase payload by 30% of its weight, keeping target waypoints the same as the training set (see Fig. 27a)

In the second test, datasets of the same 16 categories of fault scenarios are generated using an increase of 30% in its payload from 1.5kg to 1.95kg. The resulting testing accuracy falls to 80.02% (see Fig. 29). Similarly, the precision is computed to be 96.95% while the recall is computed to be much lower at 93.6%

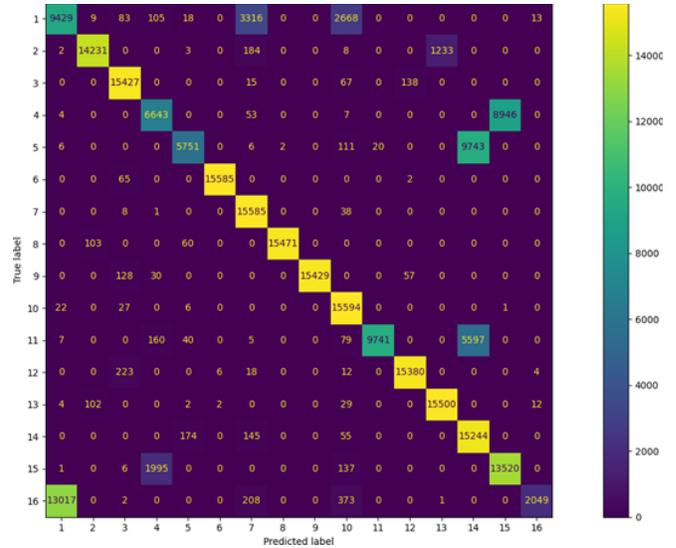

*Figure 29. Confusion Matrix of testing results (payload increased by 30%)*

Looking at the confusion matrix in Fig. 29, it is evident that the classifier struggles to differentiate between label 1 (all propellers working) and label 16 (all propellers faulty). This can be explained as a faulty propeller tend to produce lower torque and thrust; thus, the quadrotor flight path and RPM requirement resemble the case with higher payload as the propeller is required to rotate at a high speed for both scenarios.

| label | Propeller1 model | Propeller2 model | Propeller3 model | Propeller4 model |
|---|---|---|---|---|
| 1 | MP_Norm | MP_Norm | MP_Norm | MP_Norm |
| 16 | MP_Bent/MP_Crack | MP_Bent/MP_Crack | MP_Bent/MP_Crack | MP_Bent/MP_Crack |

| label | Propeller1 model | Propeller2 model | Propeller3 model | Propeller4 model |
|---|---|---|---|---|
| 4 | MP_Norm | MP_Norm | MP_Bent/MP_Crack | MP_Norm |
| 13 | MP_Bent/MP_Crack | MP_Bent/MP_Crack | MP_Norm | MP_Bent/MP_Crack |

| label | Propeller1 model | Propeller2 model | Propeller3 model | Propeller4 model |
|---|---|---|---|---|
| 3 | MP_Norm | MP_Bent/MP_Crack | MP_Norm | MP_Norm |
| 14 | MP_Bent/MP_Crack | MP_Norm | MP_Bent/MP_Crack | MP_Bent/MP_Crack |

*Figure 30. Most errors occur at these labels (too small).*

In addition, from Fig. 30, the labels show the cases where most errors occur. As this test is done with only 80 seconds worth of data per category, labels 4 and 13 tend to be miscategorized as the network has picked up the wrong information and classified the two labels based on how similar propellers 1,2,3, and 4 were. A similar case happened to labels 3 and 14 where only propeller 3 and propeller 2 are faulty respectively tend to be classified under the label where propeller 3 and propeller 2 are working.

### Testing Dataset C: Real-world flight test

In the previous section, we have shown the potential of the classification model in classifying the simulation dataset. In Dataset C, we evaluate the performance of the classifier on a real-world dataset. To gather this group of datasets, we conducted the experiment indoors as shown in the figure below. However, in this flight test, due to safety concerns, we only conducted the experiment with one faulty propeller mounted. Thus, only label 1-5 are collected for testing.

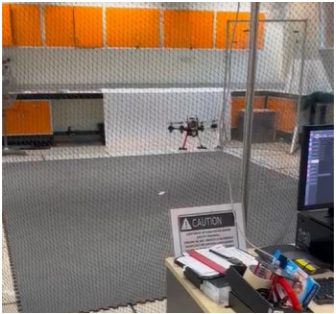

*Figure 30. Data collection indoors.*

With the CNN model trained using the adjusted simulation model mentioned in section 3.1, the model attains an average accuracy of 76.32% with a precision of 78.21% and recall of 82.39%. The reason for the drop in accuracy could be due to reasons such as overfitting and the difference in the domain the datasets are in.

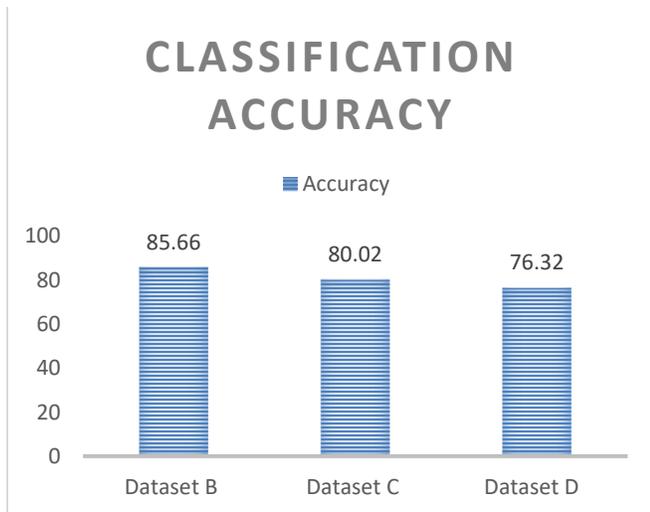

*Figure 31. Classification Accuracy of Fault Classifier*

In our future work, we are currently working on developing methods to classify the severity of the fault as well as simulate more complex environments such as the inclusion of wind. Furthermore, we understand that collecting fault data from real flights is very challenging and tedious. Thus, we are also looking into utilizing a transfer learning approach through domain adaptation to minimize the need for a real environment involving faulty UAVs. Lastly, we would also like to explore the various prognosis strategy (i.e., predict/estimate the remaining useful life of the motor/propeller system).

V. CONCLUSIONS AND FUTURE WORK

To conclude, in this paper, we have shown that it is possible to locate a faulty propeller using the target RPMs as well as the state variables as inputs. Furthermore, as all of these inputs can be computed onboard, it is possible to locate the fault while the UAV is flying. Currently, most existing approaches are based on simulation data and do not correlate well with the real world. In this paper, we have shown the effectiveness of this approach using a combination of data-driven approaches, existing dynamic models as well as adjustment to the dynamic formation to fit any real-world UAVs. We can achieve an accuracy of up to 76.32% on real-world collected data.

## References

[1] Puri, A. A Survey of Unmanned Aerial Vehicles (UAV) for Traffic Surveillance; Department of Computer Science and Engineering, University of South Florida: Tampa, FL, USA, 2005; pp. 1–29.

[2] Yao, P.; Zhu, Q.; Zhao, R. Gaussian Mixture Model and Self-Organizing Map Neural-Network-Based Coverage for Target Search in Curve-Shape Area. IEEE Trans. Cybern. 2020. [CrossRef] [PubMed]

[3] Yao, P.; Xie, Z.; Ren, P. Optimal UAV Route Planning for Coverage Search of Stationary Target in River. IEEE Trans. Control Syst. Technol. 2019, 27, 822–829. [CrossRef]

[4] Guo, Y.; Guo, J.; Liu, C.; Xiong, H.; Chai, L.; He, D. Precision Landing Test and Simulation of the Agricultural UAV on Apron. Sensors 2020, 20, 3369. [CrossRef] [PubMed]

[5] Mazzia, V.; Comba, L.; Khaliq, A.; Chiaberge, M.; Gay, P. UAV and Machine Learning Based Refinement of a Satellite-Driven Vegetation Index for Precision Agriculture. Sensors 2020, 20, 2530. [CrossRef] [PubMed]

[6] Zhi, Y.; Fu, Z.; Sun, X.; Yu, J. Security and privacy issues of UAV: A survey. Mob. Netw. Appl. 2020, 25, 95–101. [CrossRef]

[7] P. Panitsrisit and A. Ruangwiset, "Sensor system for fault detection identification and accommodation of elevator of uav," in SICE Annual Conference 2011, Sept 2011, pp. 1035–1040.

[8] Lieret, M.; Fertsch, J.; Franke, J. Fault detection for autonomous multirotors using a redundant flight control architecture. In Proceedings of the 2020 IEEE 16th International Conference on Automation Science and Engineering (CASE), Hong Kong, China, 20–21 August 2020; pp. 29–34.

[9] V. Sadhu, S. Zonouz, and D. Pompili, "On-board deep-learning-based unmanned aerial vehicle fault cause detection and identification," arXiv preprint arXiv:2005.00336, 2020.

[10] Chen F., Jiang R., Zhang K., Jiang B., Tao G. Robust Backstepping Sliding-Mode Control and Observer-Based Fault Estimation for a Quadrotor UAV. *IEEE Trans. Ind. Electron.* 2016;63:5044–5056. doi: 10.1109/TIE.2016.2552151.

[11] Avram R.C., Zhang X., Muse J. Quadrotor Actuator Fault Diagnosis and Accommodation Using Nonlinear Adaptive Estimators. *IEEE Trans. Control Syst. Technol.* 2017;**25**:2219–2226. doi: 10.1109/TCST.2016.2640941

[12] C. Rago, R. Prasanth, R. K. Mehra, and R. Fortenbaugh, "Failure detection and identification and fault tolerant control using the imm-kf with applications to the eagle-eye UAV," in Proceedings of the 37th IEEE Conference on Decision and Control (Cat. No.98CH36171), vol. 4, Dec 1998, pp. 42084213 vol.4.

[13] Zhang, H.; Gao, Q.; Pan, F. An Online Fault Diagnosis Method For Actuators Of Quadrotor UAV With Novel Configuration Based On IMM. In Proceedings of the 2020 Chinese Automation Congress (CAC), Shanghai, China, 6–8 November 2020; pp. 618–623.

[14] Zhong, Y.; Zhang, Y.; Zhang, W.; Zhan, H. Actuator and Sensor Fault Detection and Diagnosis for Unmanned Quadrotor Helicopters. IFAC-PapersOnLine **2018**, 51, 998–1003.

[15] J. Stutz, "On data-centric diagnosis of aircraft systems," IEEE Transactions on Systems, Man and Cybernetics, 2010.

[16] Guo D., Zhong M., Ji H., Liu Y., Yang R. A hybrid feature model and deep learning based fault diagnosis for unmanned aerial vehicle sensors. *Neurocomputing.* 2018;**319**:155–163. doi: 10.1016/j.neucom.2018.08.046

[17] Mallavalli, S.; Fekih, A. An SMC-based fault tolerant control design for a class of underactuated unmanned aerial vehicles. In Proceedings of the 2018 4th International Conference on Control, Automation and Robotics (ICCAR), Auckland, New Zealand, 20–23 April 2018; pp. 152–155.

[18] Liu, W.; Chen, Z.; Zheng, M. An Audio-Based Fault Diagnosis Method for Quadrotors Using Convolutional Neural Network and Transfer Learning. In Proceedings of the American Control Conference, New Orleans, LA, USA, 25–28 May 2021

[19] Chen, Y.; Zhang, C.; Zhang, Q.; Hu, X. UAV fault detection based on GA-BP neural network. In Proceedings of the 32nd Youth Academic Annual Conference of Chinese Association of Automation, Hefei, China, 19–21 May 2017.

[20] Xiao, Q.X. A Sensor Fault Diagnosis Algorithm for UAV Based on Neural Network. In Proceedings of the 2021 International.

[21] Yang P, Geng H, Wen C, Liu P. An Intelligent Quadrotor Fault Diagnosis Method Based on Novel Deep Residual Shrinkage Network. *Drones*. 2021; 5(4):133. https://doi.org/10.3390/drones5040133

[22] J. Bass. Quadrotor simulation and control (Quad_simcon). https://github.com/bobzwik/Quadrotor_SimCon, 2020.

[23] Andrew J. Barry and Russ Tedrake. High-Speed Autonomous Obstacle Avoidance with Pushbroom Stereo. PhD Thesis, March 2016.

[24] M. W. Mueller and R. D'Andrea Stability and control of a quadrocopter despite the complete loss of one, two, or three propellers 2014 IEEE International Conference on Robotics and Automation (ICRA), 2014.

[25] Barmpounakis, E.N.; Vlahogianni, E.I.; Golias, J.C. Unmanned Aerial Aircraft Systems for transportation engineering: Current practice and future challenges. Int. J. Transp. Sci. Technol. 2017, 5, 111–122. [CrossRef]

[26] B. Hou, J. Yang, P. Wang, and R. Yan, "LSTM-based auto-encoder model for ECG arrhythmias

classification," IEEE Trans. Instrum. Meas., vol. 69, no. 4, pp. 1232–1240, Apr. 2020.

[27] J. Cristian Borges Gamboa, "Deep learning for time-series analysis," 2017, arXiv:1701.01887. [Online]. Available: http://arxiv.org/abs/1701.01887

[28] T. Bresciani, "Modelling, Identification and control of a quadrotor helicopter", Master's thesis, Lund University, Sweden, 2008.